%% file: ReferenceByDescription.tex
\documentclass{llncs}
\usepackage{times}
\usepackage[labelformat=empty]{caption}
\usepackage{graphicx}
\usepackage{amsmath}
\usepackage[autostyle]{csquotes}
\usepackage[usenames,dvipsnames]{xcolor}
\usepackage{pict2e}
\usepackage{tabu}
\usepackage{wrapfig}
\usepackage{lipsum}
\usepackage{epic}
\DeclareGraphicsExtensions{.jpg, .pdf, .png}

\begin{document}

\title{Reference by Description}

\author{R.V.Guha, Vineet Gupta}
\institute{Google}

\maketitle

\begin{abstract}

Messages often refer to entities such as people, places and events.
Correct identification of the
intended reference is an essential part of  communication.
Lack of shared unique names often complicates entity reference.
Shared knowledge can be used to
construct uniquely identifying descriptive references for entities
with ambiguous names. We introduce a mathematical model for `Reference by
Description', derive results on the conditions 
under which, with high probability, programs can construct
unambiguous references to most entities in the domain of discourse
and provide empirical validation of these results.

\end{abstract}

\section{ Introduction}


Messages often need to refer to real world entities.
Communicating a reference to an entity is trivial when 
the sender and receiver share an unambiguous name for it.
However, nearly all symbols in use are ambiguous 
and could refer to multiple entities. The word 
`Lincoln' could for example, refer to the city, the president, 
or the film. In such cases, ambiguity can be resolved by 
augmenting the symbol with a unique description --- `Lincoln, the President'.
We call this {\em Reference by Description}. This leverages a combination 
of language (the possible references of `Lincoln') and 
knowledge/context (that there was only one President named Lincoln) 
that the sender and receiver share 
to unambiguously communicate a reference.

This method of disambiguation is common in
human communications. 
For example in the New York
 Times headline \cite{nytimes_jmc_obit} `John McCarthy, 
 Pioneer in Artificial Intelligence ...' the term `John McCarthy' alone is ambiguous. It
 could refer to a computer scientist, a politician or even a
 novel or film. In order to disambiguate the reference, the headline includes the
 description ``Pioneer in Artificial Intelligence''. 
An analysis, in the appendix,
of 50 articles of different genres from different newspaper/magazine
articles shows how  `Reference by Description' is ubiquitous in human communication.

The significance of correctly constructing and resolving entity references 
goes beyond human communication. 
The problem of correctly constructing and resolving entity references 
across different systems is at the core of data and application
interoperability. The following example illustrates this.

Consider an application which helps a user select and watch a movie or
TV Show.
The application has a database of movies and shows, which the user can browse
through, look at reviews from movie review sites such as
RogerEbert.com, IMDb, Rotten Tomatoes and the other (language
specific) sites. Then, it identifies a service (such as Amazon, Hulu,
Netflix, ...)  from which the movie may be
purchased or rented. Given that there are about 500,000 movies
and 3 million TV episodes, one of the most difficult parts of building
such an application is communicating references to these entities
(movies and TV shows) with these different sites. Expecting
a large number of different sites to use the same unique identifiers
for these millions of entities is unrealistic. 

Humans do not and cannot have a unique name for everything
 in the world. Yet, we communicate in our daily lives about things 
that do not have a unique name (like John McCarthy) or lack a name
(like his first car).	
 Our long term goal is to enable programs to  achieve communication
just as  effectively. 
We believe that like humans,
applications such as these  too have to use  `Reference by Description'.

The problem of entity reference is also closely related to that of
privacy preserving information sharing. When the entity about whom  
information is shared is a person, and it is done without 
the person's explicit consent (as with sharing of user profiles 
for ad targeting), it is critical this information not 
uniquely identify the person. 

We are interested in a computational model of Reference by
Description, which can answer questions such as:
What is the minimum that needs to be shared for two communicating parties
to understand each other?  How big does a description need to be?
When can we bootstrap from little to no shared language?
What is the computational cost of using 
descriptions instead of unique names?

In this paper,
we present a simple yet general model for 
`Reference by Description'. We devise measures for shared
knowledge and shared language and derive the relation between
these and the size of descriptions. From this, we answer the above
questions. We validate these answers with experiments
on a set of random graphs.

\section{Background}
 Formal study of descriptions started with Frege \cite{Frege} and
 Russell's \cite{Russell}   descriptivist theory of names,
 in which names/identity are equivalent to descriptions. 
Kripke \cite{kripke} argued against this position 
using examples where differences in domain knowledge
could yield vastly different descriptions of the same entity.  
We focus not on the philosophical underpinnings of names/identity,
but rather on enabling unambiguous communication between
software programs.

In \cite{Shannon}, the founding paper of information theory,  Shannon 
referred to this problem, saying `Frequently the messages have meaning; that 
is they refer to or are correlated according to some system 
with certain physical or conceptual entities', but he passed over it,
saying `These semantic aspects 
of communication are irrelevant to the engineering problem.'

 Computational treatments of descriptions started with linking duplicates in 
census records \cite{recordlinkage}. 
In computer science, problems in database integration, data
warehousing and data feed processing motivated the development of 
specialized algorithms for detecting duplicate items, 
 typically about people, brands and
 products. This work (\cite{Elmagarmid}, \cite{Cohen}, \cite{getoor2012entity} and
 \cite{Pasula}) has focused on identifying 
duplicates  introduced by typos, alternate punctuation, different
naming conventions, transcription errors, etc. 
 Reference resolution has also been studied in
 computational linguistics, which has developed specialized algorithms for resolving
 pronouns, anaphora, etc.

Sometimes, we can pick a representation for the domain that facilitates 
reference by description. Keys in relational databases \cite{database} are the best example of this.

The goal of privacy preserving information sharing \cite{kanonymity}
is the complement of unambiguous communication of
references, ensuring that
the information shared does not reveal the identity of the entities
referred to in the message.  \cite{dimensionality}, \cite{pii} discuss the difficulty of doing this
while \cite{searchlogs} shows how this can be done for search logs.

We have two main contributions in this paper.
Most computational treatments to date have focused on specific heuristic
algorithms for specific kinds of data. We present a general
information theoretic model for answering questions like how much
knowledge must be shared to be able to construct unambiguous references.
Second, in contrast to previous treatments which use simple
propositional / feature vector representations, we allow for richer,
relational representations.


\section{Communication model}

We extend the classical information theoretic model of communication from 
symbol sequences to sets of relations between entities. 
In our model of communication (Fig. 1):

\begin{enumerate}
\item Messages are about an underlying domain. 
A number of fields from databases and artificial intelligence to number theory
have modeled their domains as a set of entities and a set of N-tuples on these 
entities. We use this model to represent the domain.
Since arbitrary N-tuples can be constructed out of 3-tuples \cite{quine}, we  restrict
ourselves to 3-tuples, i.e., directed labeled graphs.
We will refer to the domain as {\em the graph},  and the entities as
{\em nodes}, which may be people, places, etc. or literal values such as
strings and numbers. The graph has $N$ nodes.

\item The sender and receiver each have a copy of a portion of this graph.
The arcs  that the copies have could be different, both between them
and from the underlying graph. The nodes are assumed to be the same.

\item The sender and receiver each associate a name (or other
  identifying string) 
with each arc label and zero or more names with each of the nodes in the graph. Multiple nodes 
may share the same name. Some subset of these names are shared.

\item Each message encodes a subset of the graph.  We assume that 
the sender and receiver share the grammar for encoding the message.

\item The communication is said to be successful if the receiver correctly 
identifies the nodes that the message refers to. 

\end{enumerate}

When a node's name is ambiguous,  the sender may augment the message 
with a description that uniquely identifies it.
Given a node $X$ in a graph, every subgraph that includes $X$ is a description
of $X$. If a particular subgraph cannot be mistaken for any other subgraph,
 it is a unique description for $X$. The nodes, other than
$X$, in the description are called `descriptor nodes'. In this paper,
we assume that the number arc labels is much smaller than the number
of nodes and arcs and that arc labels are unambiguous and shared.

Figures 2-4 illustrate examples of this model that 
are situated in the following context:
Sally and Dave, two researchers, are sharing information about 
students in their field. They each have some information about
students and faculty: who they work for and what universities they attend(ed). 
Fig. 1 illustrates the communication model in this context.

As the examples in Figures 2-4 illustrate,  the structure of the graph
and amount of language and knowledge shared together determine the size
of unique descriptions. 
In this paper, we are  interested in the relationship  between stochastic 
characterizations of shared knowledge, shared language and description size.
As discussed in the appendix, we can also approach this from a 
combinatorial, logical or algorithmic perspective.

In an earlier iterations of this paper \cite{G14}
we presented this model and the solution for simpler version of this problem.

\begin{figure}
\centering
\includegraphics[scale=0.35]{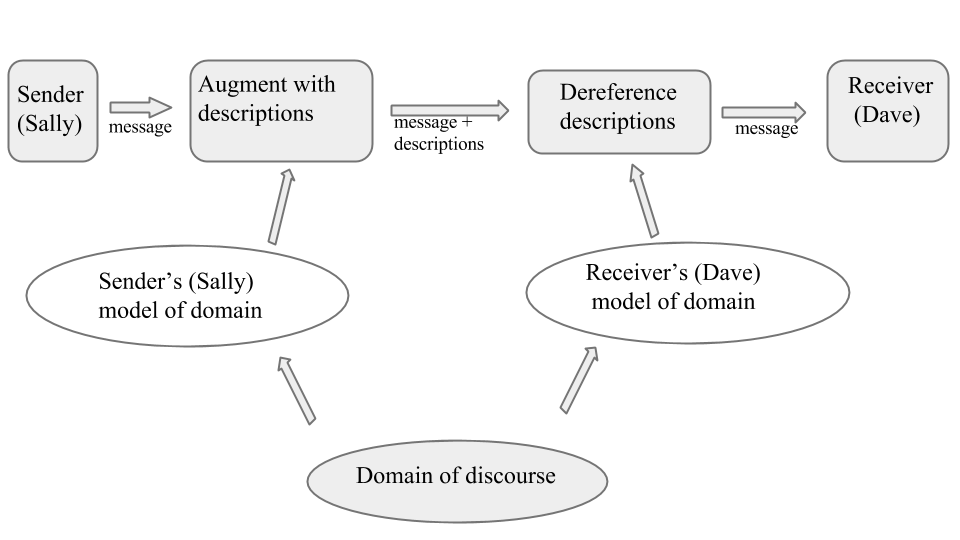}
\caption{Communication model }
\end{figure}

\begin{figure}[h!]
\centering
\includegraphics[]{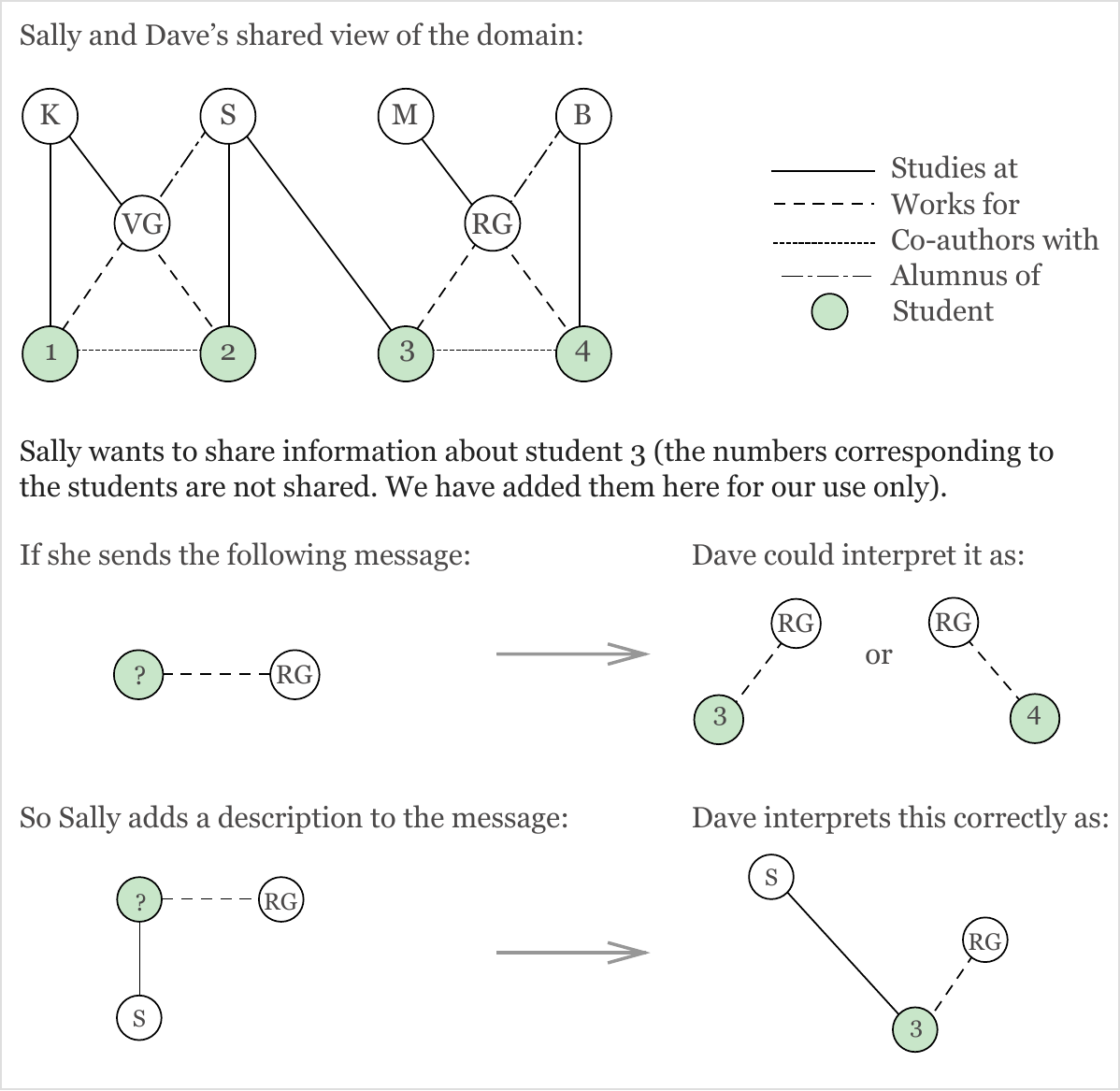}
\caption{Flat message descriptions}
\end{figure}

\begin{figure}[h!]
\centering
\includegraphics[]{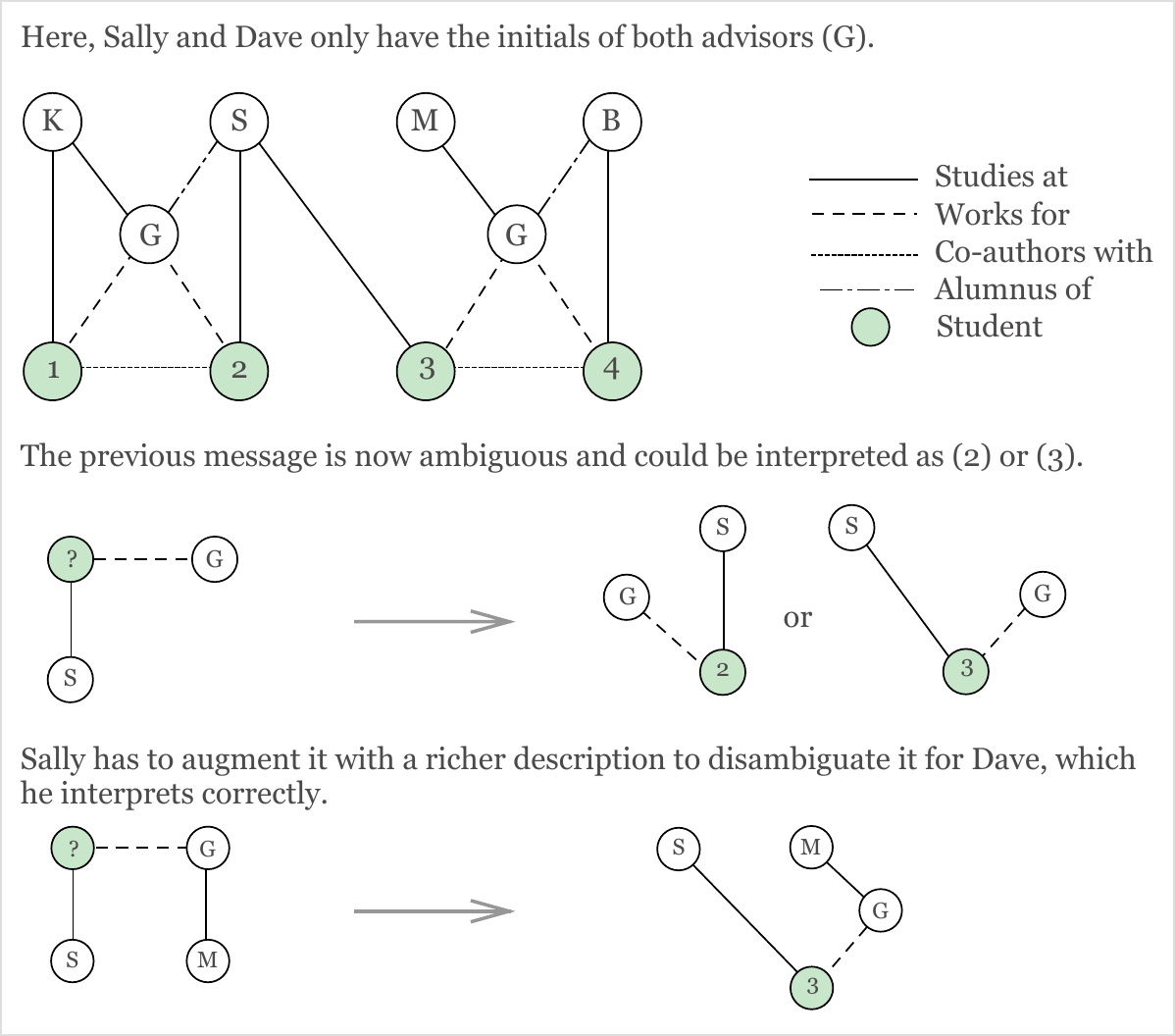}
\caption{Deep message descriptions}
\end{figure}

\begin{figure}[h!]
\centering
\includegraphics[]{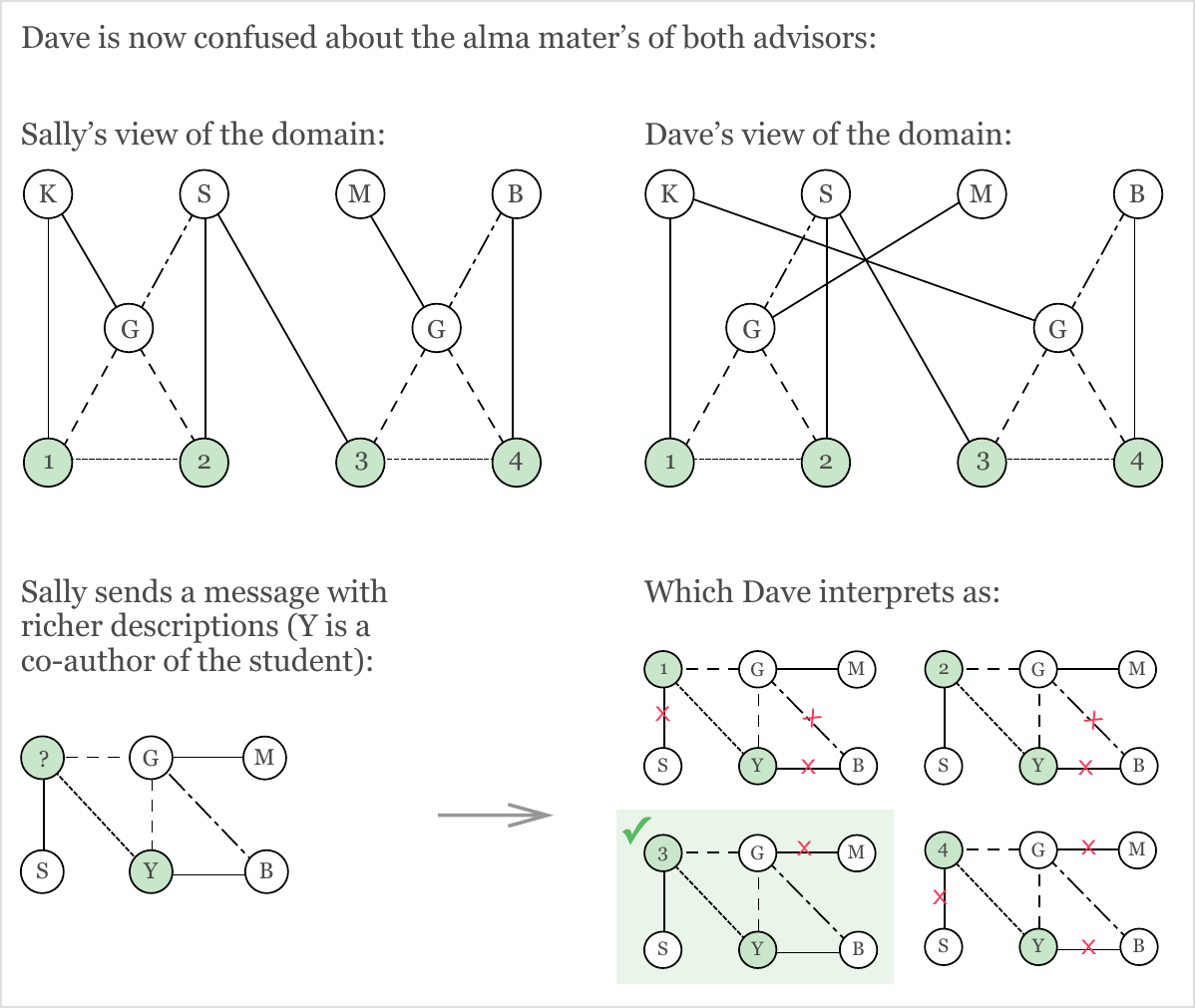}
\caption{Sally has to augment her description with annotations of the 
co-author of the student (Y). Dave can use a process similar to 
decoding Hamming codes to deduce that she is most likely describing student (3).}
\end{figure}

\section{Quantifying Sharing}

When the sender describes a node $X$  by specifying 
an arc between $X$ and a node named $N_1$, she
expects the receiver to know both which node $N_1$ refers to 
(shared language) and  which nodes have arcs going to this node 
(shared domain knowledge). We distinguish between these two.

\subsection{Shared Language / Linguistic Ambiguity}
Let $p_{ij}$ be
the probability of name $N_i$ referring to the $j\textsuperscript{th}$ object. The
{\em Ambiguity} of $N_i$ is
\begin{equation*}
A_i = \sum\limits_{j=0}^N -p_{ij} \log(p_{ij})
\end{equation*}
$A_i$ is the conditional entropy --- the entropy of the probability distribution
of over the set of entities given that the name was $N_i$. 
When there is no ambiguity in $N_i$, $A_i = 0$. 
Conversely, if $N_i$ could refer to any
node in the graph with equal probability, the ambiguity is $A_i =
\log(N)$.  
Given a set of names in a message, if we assume 
that the intended object references are independent,
the ambiguity rate associated with a sequence of names is the
average of the ambiguity of the names.
We use the Asymptotic Equipartition Property \cite{Cover} to estimate the expected 
number of candidate referents of a set of names from 
from their ambiguity. The expected number of interpretations,
of $\langle N_1 N_2 N_3 \ldots N_D\rangle$ is $2^{DA_d}$. 

Linguistic ambiguity may arise not just from a single name corresponding
to multiple entities (like the earlier example of `Lincoln') but also 
from variations of a name (such as 'Google Inc.' vs 'Google' or even 'Goggle'). Both these
can be modeled with the definition of Ambiguity discussed here.

Typically, entities with more  ambiguous names are described using entities 
less ambiguous names. We distinguish the ambiguity rate 
of the nodes being described ($A_x$) from that of the nodes used in the 
description ($A_d$).

\subsection{Shared Domain Knowledge}


Shared knowledge of the graph  enables encoding/decoding of descriptions.
Amount of sharing is not just a function of how much of the graph is shared, but the
amount of information/knowledge in the graph. 
 We want to characterize the information/knowledge
 content of the graph from the perspective of its ability to support
 distinguishing descriptions.
We introduce Salience, a measure of how much differentiation is
possible between nodes in the graph.


The classical measure for information (rate) is entropy. Let the adjancency matrix of the graph
be generated by a process with entropy $H_g$. Then the information content of a sufficiently large
{\em randomly} chosen set of $L$ arcs will be $LH_g$. However, given that the goal of our
descriptions is to distinguish a given node ($X$) from other nodes, we do not want
to randomly choose arcs involving the node. We would like to pick the arcs that are most
likely to distinguish $X$. We now introduce a quantity, which we call `Salience',
to quantify the ability of the graph to generate distinguishing descriptions.

Given a graph $G$,  there is an arc from a 
$X$ (the node being described) to each of the other nodes  in
the graph.\footnote{We label the absence of an arc between two nodes itself as a kind of arc label.}
Each of these $\langle arc-label, target \rangle$ pairs  is a `statement' about $X$.
Each description of $X$ is a subset of the statements about $X$.

Given a particular set of $D$ nodes, 
$\langle D_1, D_2, D_3, ...D_D\rangle$,
let the set of arcs from $X$ to these $D$ nodes be 
$\langle L_{XD_1}, L_{XD_2}, L_{XD_3} ... L_{XD_D} \rangle$. 
Let the probability of this sequence (or more generally, 'shape') 
occurring between these $D$ nodes and $any$
randomly chosen node in the graph be $p_i$. In Information Theoretic terms,
the information content of
 this set of arcs, or this description, is $-\log p_i$.
As $p_i$ decreases, the information content of the description
and the likelihood of the description uniquely identifying $X$ increases.
The Salience of a description its information content.

The information content of a description  grows 
with its length. The `Salience Rate' of a set of descriptions
(or a single description) is the average salience of the descriptions
in the ensemble divided by the average length (number of arcs) of the descriptions in 
the ensemble.

Consider an ensemble of size $S$ of descriptions, of average size $D$,
that hold true for a given node $X$,
containing $\gamma$ distinct shapes,
with prior probabilities of holding true for a randomly chosen other
node in the graph
of $p_1, p_2, ...p_{\gamma}$. Let the fraction of the $S$ descriptions
that have these shapes be $q_1, q_2, ...q_{\gamma}$ respectively. 
Assuming independence between the $p_i$, the Salience Rate ($F$)
of the descriptions in this ensemble is:

\begin{equation}
F = \frac{1}{D} \sum_i -q_i \log p_i
\end{equation}

Note that $p_i$ is not a distribution. The sum of
  the $p_i$'s in 
an ensemble may be less than 1. 
So, though the definition of salience bears superficial
similarity to Cross entropy and Kullback-Liebler Divergence, they are
different. 

When the description (of $X$) is constructed by randomly selecting
arcs that include $X$, the salience of the description is the entropy rate
($H_g$) of the adjacency matrix of the graph.

The salience of a description of an entity is a measure of how 
well it captures the most distinctive aspects of that entity.
The salience of an ensemble is meaningful only in the context of the
larger graph it is derived from.
Since the $p_i$ depend on the rest of the graph, the 
same ensemble, relative to a different graph might have a different salience. For example,
consider the following description: `X is a current US Senator, who studied at Harvard, ...'.
If the rest of the graph was about US senators, 
since many of the other nodes in the graph also satisfy this
description, this description has a low salience.
On the other hand, if the rest of the graph is about actors, this description has
very high salience and uniquely identifies X.

A description with a higher salience is more likely to be unique. The
sender may search through multiple candidate descriptions (of a given
salience) to find a unique description.
We are interested in an ensemble of $S$
candidate descriptions, at least one of which will 
be unique, with a probability of $1 - \epsilon$, where $\epsilon$ can
be made arbitrarity small. We will show later (equation \ref{derivation}) that for a given
$\epsilon$, as $S$ increases,  the required information
content of the description decreases, up to a certain minimum. Beyond
that, increasing $S$ does not reduce the required information content
of descriptions.
 In other  words, having more descriptions does not completely compensate for a
lack of more informative descriptions. If
we want at most a small constant number (independent of $N$) nodes to
not have unambiguous descriptions, $S \propto \log N$.

Given our interest in constructing unique descriptions,
we restrict our attention to ensembles that have the highest salience rate.
We will use the symbol $F$ for
the salience rate of an adquately large ($\log(N)$) ensemble with the highest
salience rate. Given a description of length $L$ and $D$ nodes and a salience rate of $F$,
the probability of a randomly chosen node satisfying
this description is $2^{-LF}$.

Given a set of D nodes, we have only considered the
arcs from X to these D nodes.
We can extend our treatment to include some number, say $bD$ ($b < D/2$) arcs
between the $D$ nodes in addition to the $D$ arc between 
$X$ and the $D$ nodes. The salience in such descriptions is the combination
of the salience from $X$ to the $D$ arcs plus the salience from from
the $bD$ arcs.

The salience of a graph is a measure of the underlying graph's ability to provide
unique descriptions. Differences in the view of the graph between the sender
and receiver affect how much of this ability can be used. 
 E.g., if `color' 
is one attribute of the nodes but the receiver is blind, then 
the number of distinguishable descriptions is
reduced. Some differences in the structure of the graph 
may be correlated. E.g., if the
 receiver is  color blind, some colors (such as black and white) 
may be recognized correctly  while certain other colors are indistinguishable.
We use the mutual information between the sender's and receiver's 
versions of an ensemble as the measure of
their shared knowledge. As before, consider an ensemble of size $S$ of descriptions, 
 containing $\gamma$ distinct shapes.
 Let $P(x_i)$ be the probability of the $i^{th}$ shape occurring between
 a randomly chosen node and a randomly chosen set of $D$ other nodes
 in the graph as seen by the sender.
 Given an ensemble, let $Q(x_i)$ be the probability of the $i^{th}$ shape
 occurring in the ensemble. Let $P(y_i)$ be the probability of the 
 $i^{th}$ shape occurring between
 a randomly chosen node and a randomly chosen set of $D$ other nodes
 in the graph as seen by the sender.
 Given an ensemble, let $Q(y_i)$ be the probability of the $i^{th}$ shape
 occurring in the ensemble. $P(y_i|x_i)$ is the conditional probability
 of the receiver's view having the $i^{th}$ relation between a randomly chosen node
 and $D$ descriptor nodes, given that it has this relation occurs in the
 sender's view.

 The sender chooses a description from the ensemble and sends it to the receiver
 (through a noiseless channel). The information content of the description to the
 receiver is not the same as it is to the sender. The information content of the
 description to the receiver is a function of the mutual information between
 the sender's and receiver's view of the underlying graph. Shared salience is defined as:

 \begin{equation}
 \text{Shared Salience}  = \sum_i -Q(x_i)Q(y_i|x_i) \log P(x_i)P(y_i|x_i)
 \end{equation}

 Shared salience is to salience as mutual information is to entropy. 
 If the descriptions in an ensemble are constructed by randomly choosing
statements or if the descriptions become very large,
shared salience tends to the mutual information.

\subsection{Salience of Random Graphs}

Given our interest in probabilistic estimates on the 
sharing required, we focus on graphs generated by stochastic
processes, i.e., Random Graphs. The most commonly studied
Random Graph model is that proposed by Erdos and Renyi \cite{erdos}.
The Erdos-Renyi random graph $G(N,p)$ has $N$ nodes where the
probability of a randomly chosen pair of nodes being connected is $p$.
The Salience Rate of a sufficiently large $G(N,p)$ graph
is $-\log(p)$ (or $-\log(1-p)$, whichever is larger).

More recent work on stochastic graph models 
has tried to capture some of the phenomenon found in real world
graphs. Watts, Newman, et. al. \cite{watts},
\cite{newman1999} study small world graphs, where there are a large number
of localized clusters and yet, most nodes can
be reach from every other node in a small number of hops. This phenomenon
is often observed in social network graphs. 
`Knowledge graphs', such as DBPedia \cite{dbpedia} and Freebase \cite{freebase}, that
represent relations between people, places, events, etc., tend to exhibit
complex dependencies between the different arcs in/out of a node. Learning
probabilistic models \cite{daphne} of such dependencies is an active area
of research.

$G(N,p)$ graphs assume independence between arcs, i.e., the probability
of an arc $L_i$ occurring between two nodes is independent of all other
arcs in the graph. Clustering and the kind of phenomenon found in 
 knowledge graphs can be modeled
by discharging this independence assumption and replacing it with 
appropriate conditional probabilities. For example, the clustering
phenomenon occurs when the probability of two nodes $A$ and $C$ being
connected (by some arc) is higher if there is a third node $B$ that is
connected to both of them.

When the graph is generated by a first order Markov process, i.e.,
the probability of an arc appearing between two nodes is
independent of other arcs in the graph, as in $G(N,p)$, 
calculating $F$ is simple. For more complex graphs
where there are conditional dependencies between the arcs in/out from
a node, we need to model the graph generating process as a 
second or even higher order Markov processes to compute $F$.

\subsection{Description Shapes}

The number of nodes ($D$) in a description
of length ($L$) depends on its shape. The decoding complexity of a
description is a function of both its size and its shape.
Flat descriptions, which are the easiest to decode only 
use arcs between the node being described and the nodes 
in the description. E.g., Jim, who lives in Palo Alto, CA, age 56 yrs,
works for Stanford, studied at UC Berkeley, married to Jane.
The description consists of the arcs from node being described, Jim,
to the descriptor nodes, Palo Alto CA, 56 yrs, Stanford, UC Berkeley
and Jane.
The length ($L$) flat descriptions, i.e., the number of arcs included, is the
same as the number of nodes ($D$), i.e., $L = D$. 
They have $O(aD)$ decoding complexity, where $a$ is the average degree
of each node.

\input{description_figs.tex}

Flat descriptions are most effective when the descriptors
have low ambiguity and do not require disambiguation themselves.
In the previous description of Jim, the terms `Palo Alto, CA', `Stanford'
and `UC Berkeley' are relatively unambiguous. 
Most descriptions in daily communication are relatively flat. 
When  low ambiguity descriptor terms are not available, the descriptor terms
might need to be disambiguated themselves. In such
cases adding `depth' to the description is helpful. 

In their most general form, deep 
descriptions do not impose any constraints on the
shape. E.g., Jim, who is married to Jane who went
to school with Jim's sister and whose sister's child goes to the same school
that Jim's child goes to. This description includes
a number of links between the descriptor nodes (Jane, Jane's school,
Jim's sister, etc.). The goal is  capture
some unique  set of  relationships that serve to unambiguously
identify the node being described.
Decoding deep descriptions involves
solving a subgraph isomorphism problem and is NP-complete.
These descriptions have length up to $L = D^2/2$ and have $O(N^D)$ decoding complexity.

A trade off between decoding complexity and expressiveness can
be achieved with by constraining the arcs (between descriptor nodes) that
are included in the description. More specifically, the $D$ nodes
in the description can be arranged into square blocks
where only the arcs within each block are included in the description.
We assume that there are $Db$ links between the descriptor nodes
giving us $L = D(b+1)$. 
This is illustrated in Fig. 1, with the label 'intermediate'.
When $b = D/2$, these reduce to the 
general form of deep descriptions.

\section{Description size for unambiguous reference}

{\bf Problem Statement:} The sender is trying to communicate a message that mentions 
a large number of
randomly chosen entities, whose average ambiguity is $A_x$.   The overall graph has $N$ nodes.
Each entity in the message has a description, involving
on average, $D$ descriptor nodes,  whose ambiguity rate is $A_d$. 
The description includes $bD$ ($b \le D/2$) arcs between the descriptor nodes,
which may be used to reduce the ambiguity in the descriptor nodes themselves, if any.
We are interested in the average number of arcs and nodes
required in the description.

In the most general terms, the ambiguity resolved by a description
is less than or equal to its information content. 
More specifically, if $F$ is the salience rate of the graph, 
under the assumption of uniform salience rate, we have:

\begin{equation}
\label{basic}
D = \frac{A_x}{F - \max(0, A_d - bF)}
\end{equation}





\noindent Equation \ref{basic} covers a range
of communication scenarios, some of which we discuss now.
Proofs and empirical validation of equation \ref{basic}
and its  application to these scenarios 
are in the appendix
We first examine the impact of the structure of the description,
which affects the computational cost of
constructing and decoding it.

\subsection{Flat Descriptions}
Flat descriptions (Fig. 2), which are the easiest to decode, only 
use arcs between the node being described and the descriptor nodes.
They can be decoded in $O(aD)$, where $a$ is the average degree of a node.
For flat descriptions, $b = 0$, giving,
\vspace{-.08in}
\begin{eqnarray}
\label{short-flat}
D = \frac{A_x}{F - A_d} & \text{\ \ \ Flat descriptions}
\end{eqnarray}

Flat descriptions are very easy to decode, but require
relatively unambiguous descriptor nodes, i.e., $F \gg A_d$.
Most descriptions in human communication fall into this category.

\subsection{Deep Descriptions}
If the descriptor nodes themselves are very ambiguous ($F - A_d$ is
small),  the ambiguity of the descriptor nodes can
be reduced by adding $bD$ arcs between them. 
If the descriptor nodes are considerably less ambiguous
than the node being described ($A_d < A_x/2$), all ambiguity in
the descriptor nodes can be eliminated by including $A_d/F$ links between them,
giving us:

\begin{eqnarray}
\label{short-deep}
D = \frac{A_x}{F} & \text{\ \ \ Deep  descriptions}
\end{eqnarray}

Fig. 3 and 4 are examples of deep descriptions.
Deep descriptions are more expressive and can be used
even when the descriptor nodes are highly ambiguous.
However, decoding deep descriptions involves solving a subgraph isomorphism 
problem and have $O(N^D)$ decoding complexity.


\subsection{Purely Structural Descriptions}

When the sender and receiver don't share any linguistic knowledge, 
all nodes are maximally ambiguious ($A_x = A_d = \log N$).
We have to rely purely on the structure of the graph. We have:

\begin{eqnarray}
\label{sld}
D = 2\log(N)/F & \hspace{.1in}  &\text{Purely structural descriptions}
\end{eqnarray}

By using detailed descriptions that include
multiple attributes of each of the descriptor nodes, we can bootstrap
communication even when there is almost no shared language.

\subsection{Limiting Sender Computation}

The sender may not be able to search through multiple
candidate descriptions, checking for uniqueness. We are interested
in $D$ such that every candidate description of size $D$ and salience
rate $F$ is very likely unique. Assuming unambiguous descriptor nodes, we have: 
\vspace{-0.04in}
\begin{eqnarray}
\label{landmark-basic}
 D = \frac{\log(N) + A_x}{F}  & \text{\ \ \ Flat Landmark descriptions}
\end{eqnarray}

When the descriptions are constructed by randomly choosing facts
about the entity, the salience rate is equal to the entropy of the adjacency
matrix of the graph, $H_g$. In this case, the the nodes can use the same
set of descriptor nodes, whence the name `landmark descriptions'.

\subsection{Language vs Knowledge + Computation trade off}
 Consider a node with no name ($A_x = \log(N)$). Given a set
of candidate descriptions with salience rate $F$, we consider two
kinds of descriptions which are at opposite ends of the spectrum
in the use of language vs knowledge.
We could use purely structural descriptions (eq \ref{sld}), which use 
no shared language.
We could also, use a flat landmark description (eq. \ref{landmark-basic})
which makes much greater use shared language and ignores most of 
shared graph structure/knowledge. Though the number of nodes
$D = 2\log(N)/F$ is the same in both, flat descriptions are of length $O(D)$, require no
computation to generate and can be interpreted
in time $O(aD)$. The former, in contrast are of length $O(D^2)$. 
Further, since generating and interpreting
them involves solving a subgraph isomorphism problem, they may require
$O(N^{2\log(N)/F})$ time to interpret. 
This contrast illustrates tradeoff between shared language, shared domain 
knowledge and computational cost.  We can overcome the lack
of shared language by using shared knowledge, but only at the cost of
exponential computation.

\subsection{Minimum Sharing Required}
When there are relatively unambiguous descriptor nodes available, 
$A_x/F$, the minimum size of the description for $X$, is a measure
of the  difficulty of communicating a reference
to $X$. It can be high either because $X$ is very ambiguous ($A_x \to \log(N)$) 
and/or because very little unique is known about it ($F \to 0$). 
When $A_x/F \ge N$ it is not possible to communicate a reference to $X$.  
 As the ambiguity of the descriptor nodes increases, domain knowledge
has to play a greater role in disambiguation. In the limit, when
there are no names, we have to use purely structural descriptions.
In this case, $2\log(N)/F$ has to be less than $N$.


\subsection{Non-identifying descriptions}
We are interested in comparing the number of statements
that can be made about an entity, while still keeping it
indistinguishable from $K$ other entities \cite{kanonymity},
with the number of statements required to uniquely identify it.
For this comparison to be meaningful, in both cases, we use
statements with the same salience rate ($F$).
Since the purpose is to hide the identity of the entity,
its name is not included in the description, i.e., $A_x = \log(N)$. 
For flat descriptions we have:

\begin{equation}
\label{nonidentifying}
 D \leq \frac{\log(N) - \log(K)}{F-A_d}
\end{equation}

Comparing this to equation \ref{short-flat} (with $A_x = \log(N)$), we
see that there is only a small size difference ($\log(K)/(F-A_d)$) between K-Anonymous
descriptions and the shortest unique description. This is because
of the phase change (discussed in the appendix), wherein at around
$D = A_x/(F - A_d)$, the probability of finding a unique description
of size $D$ abruptly goes from $\approx 0$ to $\approx 1$.
Though most descriptions of size $A_x/(F - A_d)$ are
not unique, for every node, there is at least one such description that is unique.

Given the statistical nature of equation \ref{nonidentifying}, it is  
a neccessary, but not sufficient condition for privacy.
Given a large set of entity descriptions, if the average
size of description is close to or larger than this limit,
then, with high probability, at least some of the entities 
have been uniquely identified.

\section{Conclusion}

 As Shannon \cite{Shannon} alluded to, communication is not just
correctly transmitting a symbol sequence, but
also understanding what these symbols denote. 

Even when the symbols are ambiguous,
using descriptions, the sender can unambiguously communicate which 
entities the symbols refer to.
We introduced a model for `Reference by Description' and
show how the size of the description goes up as the amount
of shared knowledge, both linguistic and domain, goes down.
We showed how unambiguous references can be constructed from purely
ambiguous terms, at the cost of added computation. 
The framework in this paper opens many 
directions for next steps:

\begin{itemize}
\item Our model makes a number of simplifying assumptions.
It assumes that the sender
has knowledge of what the receiver knows.
An example of this assumption breaking down
is when two strangers speaking different languages have to communicate.
It often involves a protocol of pointing to something and uttering
its name in order to both establish some shared names and to
understand what the other knows.
A related problem appears in the case of broadcast communication, where different
receivers may have different levels of knowledge, some of which is
unknown to the sender. 
A richer model, that incorporates a  probabilistic
characterization of not just the domain, but also of the receiver's
knowledge, would be a big step towards capturing these phenomena.

\item We have assumed large graphs and long messages. In practice,
context is used to circumscribe the graph. Understanding the
relation between context and descriptions would be very interesting.

\item Though our communication model makes no assumptions about the
graph, the simple form of the results presented here arise out of
assumptions about ergodicity and uniformity of salience rate (which
are analogous to those made in \cite{Shannon}). Versions of these
results that don't make these assumptions would be useful.

\item Though we have touched briefly on practical applications
of our model, much work remains to be done.
The first task is the development of algorithms for
constructing unique descriptions. 

\end{itemize}

\section*{Acknowledgments}
The first author thanks Phokion Kolaitis and Andrew Tompkins for providing
a home in IBM research to start this work and  Bill Coughran at Google
to complete it. Carolyn Au did the figures.
Carolyn Au, Vint Cerf, Madeleine Clark, Evgeniy Gabrilovich,
 Neel Guha, Sreya Guha, Asha Guha, Joe Halpern, Maggie Johnson, Brendan Juba, 
  Arun Majumdar,  Peter Norvig, Mukund Sundarajan and Alfred Spector
 provided feedback on drafts of this paper.

\bibliographystyle{abbrv}
\bibliography{cr}

\newpage
\appendix

{\bf Appendices} \\
We have the following appendices:

\begin{description}

\item[Appendix A:] Derivation of results and various special cases
\item[Appendix B:] Empirical validation of results on random graphs
\item[Appendix C:] Studies on usage of descriptive references on a set
of newspaper/magazine articles
\item[Appendix D:] Alternate problem formulations
\end{description}





\section{Derivation of Results}

\subsection{Problem Statement}
We are given a large graph $G$ with $N$ nodes and a
message, which is a subgraph of $G$. The message
contains a number of randomly chosen nodes. We
construct  descriptions for each of the nodes ($X$) in this
subgraph so that it is uniquely identified.
We are interested in a stochastic characterization of the relationship
between the amount of shared domain knowledge, shared language,
the number of nodes in the description ($D$) and the number of arcs ($L$)
in the description. 

 For expository reasons,
we go through the
derivation for the simpler case, where there is no
ambiguity in the descriptor nodes. We then extend this
proof to the case where the descriptor nodes themselves
may be ambiguous.

We model the graph corresponding to the domain of discourse, and the 
message being transmitted, as being generated by a stochastic processes.
We carry over the assumption from Shannon \cite{Shannon} 
and information theory \cite{Cover} that sequences of symbols (in
our case, entries in  the adjacency matrix) are generated
by an ergodic process.



\subsection{Unambiguous Descriptor Nodes}

Consider a node $X$ in the message, which has an ambiguity 
of $A_x$\footnote{More precisely, we let the 
average ambiguity rate of the nodes in the message be
$A_x$. Henceforth, even though we are dealing with average value of
properties of  entities in the message, for the sake of clarity in the
presentation,
we will treat the corresponding properties of the node $X$  
and its descriptor nodes as proxies for
the average.}. 
The description for $X$ involves $D$ unambiguous descriptor nodes.
There are a  number of possible configurations of $D$ arcs from $X$ to the descriptor
nodes. Let us call these $\langle C_{XD1}, C_{XD2}, C_{XD3}, ...\rangle$. 
Let the probability of the $i^{th}$ of these occuring between a
randomly chosen node a set of $D$ descriptor nodes be $p_i$.

The sender looks through an ensemble of $S$ descriptions, i.e., 
$S$ possible combinations of $D$ descriptor nodes
in search of a unique description for $X$. 
Let the probability a randomly chosen element of this ensemble 
of these descriptions having the configuration $C_{XDi}$ be $q_i$.

There are $2^{A_x} -1$ nodes that might be mistaken for $X$. 
Consider one particular element in the ensemble that has the configuration
$C_{XDi}$ with $X$. It  provides a disambiguating description
for $X$ if none of the $2^{A_x} -1$ nodes that we are trying to distinguish  
$X$ from is also in the configuration $C_{XDi}$ with this set of descriptor
  nodes. The probability of this is 

\begin{equation}
(1 - p_i)^{2^{A_x} - 1}
\end{equation}

\noindent There are $S$ such sets of descriptors.
We want the  probability of none of these $S$ descriptions being unique to
be less than $\epsilon$. Assuming independence between the
descriptions in the ensemble, the probability of this is:

\begin{equation}
\label{bb}
\prod_{D_i\in S} (1 -  (1 - p_i)^{2^{A_x} - 1}) \le \epsilon
\end{equation}

This product is over the candidate descriptions in the ensemble.
Since we are interested in descriptions that are unlikely to be satisfied
by other nodes, we can restrict our attention to the case where $p_i$ is
very small, or more specifically, $(2^{A_x}-1)p_i \ll 1$. This allows
us to use the binomial approximation, using which we get:

\begin{equation}
\prod_{D_i\in S} (1 - (1 - p_i (2^{A_x} - 1))) \le \epsilon
\end{equation}

\noindent Assuming $2^{A_x} \gg 1$ 
\begin{equation}
\prod_{D_i\in S} p_i 2^{A_x}   \le \epsilon
\end{equation}

\noindent Taking logs, 

\begin{equation}
\label{e8}
\sum_{D_i\in S} \log p_i  + SA_x \le \log(\epsilon)
\end{equation}

For sufficiently large $S$, the
different description configurations will occur in proportion to their
likelihood, i.e. the number of times shape $C_{XDi}$ occurs in the set $S$
is approximately $Sq_i$. So, $\sum_{D_i\in S} \log p_i$ can be rewritten as,
$\sum_{j} Sq_j \log p_j$ where $j$ ranges over all the possible description
configurations.

\begin{equation}
\sum_{D_i\in S} \log p_i = \sum_{j} Sq_j\log p_j = -SF_D
\end{equation}
\noindent where $F_D$ is the average salience of the descriptions in  the ensemble.

\begin{equation}
\label{derivation}
F_D  \ge A_x - \frac{\log(\epsilon)}{S}
\end{equation}

From this, we see that the impact of the size of $S$ on the average
$F_D$ required is a function of $\epsilon$. Increasing $S$ beyond
the stage where $\frac{\log(\epsilon)}{S}$ is sufficiently small does
not provide additional benefit.

If we want at most a small constant (say one) node to not have
a unique description, $\epsilon = 1/N$, in which case we get,

\begin{equation}
F_D  = A_x - \frac{\log(N)}{S}
\end{equation}

\noindent If S is sufficiently large so that $\frac{\log(N)}{S}$
is negligible, we get, 

\begin{equation}
F_D  = A_x 
\end{equation}

\noindent If the salience rate of the ensemble is $F$, i.e., $FD =
F_D$,  where $D$ is the average length of the description in the
ensemble, we have

\begin{equation}
\label{simple}
D  = A_x/F
\end{equation}

This gives us an upper bound on $D$. 
 We now show that this is also a lower bound (under the
assumption, $A_x \gg 0$). To simplify, we demonstrate the proof for
the average case, where $F_D = DF$. We are interested
in large graphs where the number of ambiguous nodes does
not grow with the size of the graph. So we let $\epsilon = 1/N$. We can rewrite
equation \ref{bb} as:

\begin{equation}
2^{(A_x - DF)S} = 2^{(A_x +\frac{\log(N)}{S}- DF)S} 
\end{equation}

\noindent This is the number of ambiguous nodes. Let

\begin{equation}
D = \frac{(1 + \delta)(A_x + \frac{\log(N)}{S})}{F}
\end{equation}

\noindent where $\delta$ can be positive or negative. We get

\begin{equation}
2^{\delta (A_x + log(N)/S)S} = \text{Number of ambiguous nodes}
\end{equation}

\noindent Clearly, for large $N$, the only way the
number of ambiguous nodes does not grow with $N$ is if $\delta \le 0$, 
showing that equation \ref{basic} is a lower bound as well. 

A variant of this derivation shows the sensitivity of the whether
a set of descriptions are unambiguous to the description size.
Using the earlier approximations, we get the probability $p_u$ of a
node having a uniquely identifying description of size $D$ as:

\begin{equation}
p_u = (1 - (1 - 2^{-FD})^{2^{A_x}} )^S
\end{equation}

\begin{equation}
p_u = 2^{(A_x  - FD)S}
\end{equation}

\noindent Let $D = \frac{A_x}{F} + \delta$, where $\delta$ can be
positive or negative. This gives us

\begin{equation}
p_u = 2^{\delta S}
\end{equation}

\noindent  Given the size of $S$ ($O(\log N)$),  in the exponent, for 
large $N$, it is easy to see
how as $\delta$ goes from negative to positive, at around 
$\delta = 0$, there is a `phase change' and $p_u$ abruptly goes from being $p_u \approx 0$ to
$p_u \approx 1$. So, for a certain $D$ which is just less than given by
equation \ref{basic} almost none of the nodes have
unique descriptions. When the description size reaches that given by
equation \ref{basic} there is an abrupt change and almost all nodes
have unique descriptions.

If the shared salience between the sender and receiver is $M$, then,
by using arguments identical to those in \cite{Shannon}, we have

\begin{equation}
D  = A_x/M
\end{equation}

Note again that this is the upper
bound on the average number of descriptor nodes for the entities in a
sufficiently long message. When $A_x$ is low, description sizes will
be small and individual nodes in the message may have
shorter or longer descriptions. However, this bound still applies to
the average description length for a large set of nodes.

\subsection{Ambiguous Descriptor Nodes}

We now consider the case where the descriptor nodes themselves are
ambiguous. Let the ambiguity rate of the descriptor nodes be $A_d$. 
The description also includes $bD$ ($0 \le b \le D/2$) 
arcs between the descriptor nodes. The role of these $bD$ arcs
is to reduce the ambiguity of the descriptor nodes.

There are a number of possible configurations of $bD$ arcs between
a randomly chosen set of $D$ candidate descriptor nodes. 
Let us call these $\langle C_{bD1}, C_{bD2}, C_{bD3} ...\rangle$
Let the  probability of the $i^{th}$ of these occurring amongst
a randomly chosen set of $D$ descriptor nodes be $q_i$\footnote{Ambiguity
amongst the descriptor nodes themselves might lead to some of these
configuration of arcs might being automorphic to others. In this
paper, we ignore this.}. 

The sender looks through  $S$ possible combinations of $D$ descriptor nodes
in search of a unique description for $X$. 
Consider one particular set $i$ of $D$ descriptor nodes for  $X$. Let
the configuration of the arcs between $X$ and the $D$ nodes be
$C_{XDi}$ and the configuration of the $bD$ arcs between the $D$ nodes
be $C_{bDi}$.

There are $2^{DA_d}$ sets of $D$ nodes which have the same names as 
these descriptor nodes. Only one of these is the intended set of
descriptor nodes.
One of  the other  $2^{DA_d}-1$ sets of descriptor nodes can be mistaken for the
intended descriptor nodes only if the $bD$ arcs between them also
has the configuration $C_{bDi}$, the probability of which is $q_i$.
Similarly, there are $2^{A_x} -1$ nodes that might be mistaken for
 $X$. The probability of one of these having the configuration
 $C_{XDi}$ with a set of $D$ descriptor nodes is $p_i$.
This set of descriptor nodes provides a disambiguating description
for $X$ if, 
\begin{enumerate}
\item None of the $2^{A_x} -1$ nodes that we are trying to distinguish
  $X$ from is in the configuration $C_{XDi}$ with this set of descriptor
  nodes AND
\item None of the $2^{A_x} -1$ nodes that we are trying to distinguish
  $X$ from is in the configuration $C_{XDi}$ with one of the set
  of nodes that the descriptor nodes could be mistaken for.
\end{enumerate}

\noindent The probability of this set of descriptor nodes 
providing a unique description for $X$ is:

\begin{equation}
 (1 - p_i)^{2^{A_x} - 1}(1 - p_iq_i)^{(2^{A_x}-1)(2^{DA_d} - 1)} 
\end{equation}

\noindent As before, assume that $2^{A_x} \gg 1$. To simplify the
analysis, we only consider the case where the ambiguity in the
descriptor nodes is not insignificant, i.e., $2^{DA_d} \gg 1$. With these
assumptions, we get:

\begin{equation}
 (1 - p_i)^{2^{A_x}}(1 - p_iq_i)^{2^{A_x + DA_d}} 
\end{equation}

\noindent There are $S$ such sets of descriptors.
The probability of none of these $S$ descriptions being unique should
be less than $\epsilon$.

\begin{equation}
\label{bb}
\prod_{D_i\in S} (1 - (1 - p_i)^{2^{A_x}}(1 - p_iq_i)^{2^{A_x + DA_d}}) < \epsilon
\end{equation}

\noindent Using the binomial approximation (as before, we assume that  the
number of descriptions is large and hence $p_i$ and $p_iq_i$ are very small), 

\begin{equation}
\prod_{D_i\in S} (1 - (1 - p_i 2^{A_x})(1 - p_iq_i2^{A_x +DA_d})) < \epsilon
\end{equation}

\noindent Multiplying out, and ignoring terms with higher
powers of $p_i$ and $q_i$,

\begin{equation}
 \prod_{D_i\in S} p_i2^{A_x}\bigl(1 + q_i2^{DA_d}\bigr) < \epsilon
\end{equation}

\noindent Taking logs, 

\begin{equation}
\label{e8}
\sum_{D_i\in S} \log p_i  + SA_x + \sum_{D_i\in S}\log (1 + q_i2^{DA_d}) < \epsilon
\end{equation}

\noindent $q_i$ is the probability of the $i^{th}$ shape in the ensemble 
occurring between a randomly chosen set of $D$ nodes and is equal to $2^{-bDF}$, 
where $F$ is the salience rate for the ensemble with respect to the graph. So,

\begin{equation}
\sum_{D_i\in S} \log p_i  + SA_x + \sum_{D_i\in S}\log (1 + 2^{DA_d -bDF}) < \epsilon
\end{equation}

Assume $\log (1 + 2^{DA_d -bDF}) \approx D(A_d - bF), bF \le A_d$. If
$bF > A_d$, $\log (1 + 2^{DA_d -bDF}) \approx 0$. So, 
$\log (1 + 2^{DA_d -bDF}) \approx Dmax(0, A_d - bF)$.
Letting $p_i = 2^{-DF}$ as before, we get:

\begin{equation}
\label{basic}
D \approx \frac{\log(N) / S + A_x}{F - max(0, A_d - bF)}
\end{equation}

\subsection{Searching through candidate descriptions}

Description size (and hence decoding cost) is influenced by $S$, the
number of potential descriptions the sender can search through. 
$S=1$ corresponds to the case where $D$ is sufficiently large, so that 
any selection of $D$ nodes will likely
form a unique shape. This minimizes the sender's computation at the
expense of description length and receivers compute cost.

The $D$ nodes can be selected such that the same $D$ nodes are used to
describe all the remaining $N-D$ nodes
or each node can use a different set of $D$ nodes to describe it.
We call these `landmarks' nodes and the associated descriptions
are called `landmark descriptions'. From eq. (\ref{basic}), we have:

\begin{equation}
\label{landmark-basic}
D = \frac{\log(N) + A_x}{F - max(0, A_d - bF)}
\end{equation}

Intuitively, the size of $D$ given by equation \ref{landmark-basic}
answers the following question: how many randomly chosen facts, 
at the salience rate $F$, about
an node does one have to specify to uniquely identify that node,
with high probability. Note that in this case, the sender is not
looking at the other nodes in the domain to see if the
description is unique. If the description is longer that the size
given by equation \ref{landmark-basic}, it is very likely unique.

Remember that if the statements are chosen at random from the graph, the salience
rate is $H_g$. So, restricting ourselves to flat descriptions composed
of randomly chosen statements about the object, we have:

\begin{equation}
\label{landmark-random}
D = \frac{\log(N) + A_x}{H_g - A_d}
\end{equation}

One interesting case of this is where the `randomly' chosen nodes (in
terms of which $X$ is described) is the same for all $X$. Such a set
of descriptor nodes serve as 'landmarks' in terms of which all other
nodes are described. If the landmark nodes are unambiguous and the
$X$ have no name, we have the special case where:

\begin{equation}
\label{landmark-random-simple}
D = \frac{2\log(N)}{H_g}
\end{equation}

At the other extreme, the sender could search through enough 
descriptions to find the smallest set of descriptor nodes for uniquely
identifying $X$. For each description, the sender checks to see if the
description is indeed unique.
It is enough for the sender to search through $S$ randomly chosen
descriptions such
that $\frac{\log N}{S} \approx C$, where $C$ is a small constant (which can
be ignored). In practice, by going through candidate descriptions 
in something better than random order, far fewer than $O(\log(N))$ 
descriptions need to be considered.  In this case:

\begin{equation}
\label{adaptive-basic}
D = \frac{A_x}{F - max(0, A_d - bF)}
\end{equation}

\subsection{Flat Descriptions}
\label{flat}

The shape of the description influences decoding complexity. 
Flat descriptions, which are the easiest to decode, only 
use arcs between the node being described and the nodes 
in the description --- no arcs between other nodes in the description
are considered.  Thus for these $b = 0$.

\begin{eqnarray}
\label{adaptive-flat}
D = \frac{A_x}{F - A_d} & \text{\ \ \ Flat descriptions}
\end{eqnarray}

\noindent In the case where we do not have a name for the node being
described, we get:

\begin{eqnarray}
\label{adaptive-flat-anon}
D = \frac{\log N}{F - A_d} & \text{\ \ \ Flat descriptions}
\end{eqnarray}

\indent As expected, flat descriptions are longer when the sender can not search
through multiple candidates.

\begin{eqnarray}
\label{landmark-flat}
D = \frac{\log(N) + A_x}{F - A_d}  & \text{\ \ \ Flat landmark
  descriptions}
\end{eqnarray}

\noindent If $F < A_d$, we cannot use flat descriptions. 





\subsection{Deep Descriptions}

When $A_d > 0$, the number of nodes in the description may be reduced
by using deep descriptions.
In deep descriptions, in addition to the arcs
between the descriptor nodes and $X$,
arcs between the $D$ nodes may also be included in the description. We
include $bD$ arcs between the descriptor nodes in the description.

We can restrict the search for descriptions (i.e., $S = 1$), giving 
us deep landmark descriptions. If $b \leq \frac{A_d}{F}$, we have:

\begin{eqnarray}
\label{landmark-deep}
D = \frac{\log(N) + A_x}{(b+1)F -A_d} & \text{\ \ \ Deep Landmark Descriptions}
\end{eqnarray}

Note that if $(b+1)F < A_d$, then communication will not be possible.
When the descriptor nodes are unambiguous, adding depth does not provide any utility.
For sufficiently large $S$,

\begin{eqnarray}
\label{da}
D = \frac{A_x}{(b+1)F -A_d} & \text{\ \ \ Deep Descriptions}
\end{eqnarray}

When $A_d < 2A_x$ and $A_d/F < D/2$ and we can eliminate ambiguity in the descriptor
nodes without increasing $D$, giving us:

\begin{eqnarray}
\label{short-deep}
D = \frac{ A_x}{F} & \text{Deep descriptions}
\end{eqnarray}

\noindent Given a particular $\langle F, A_d, A_x \rangle$,  deep 
descriptions have the fewest nodes. As $F$ decreases or $A_x$ (or
$A_d$) increases, we need bigger descriptions.

\subsection{Purely Structural Descriptions}

 For purely structural descriptions ({\em i.e.} no names), $A_x = A_d
 = \log N$.  Allowing $b = D/2$ in equation \ref{da}:

\begin{equation}
D = \frac{\log(N)}{(D/2+1)F -\log(N)} \approx \frac{\log(N)}{DF/2 -\log(N)}
\end{equation}

\begin{equation}
D^2F = 2D(1 + \log(N)) \approx 2D\log(N)
\end{equation}

\noindent From which we get:

\begin{equation}
D = \frac{2\log(N)}{F}
\end{equation}

\subsection{Message Composition / Self Describing Messages}

\label{coherent}
 We have allowed for the entities in a message to be randomly chosen.
The entities in the message may or may not have 
relations between them. For example, if the message is a set of
census records or entries from a phone book, the different entities in the message
will likely not be part of each other's short descriptions. 
In contrast, in a message like a news article, the entities that
appear do so because of the relations between them and can
be expected to appear in each other's descriptions. 

We call  messages, where all the descriptors for the entities
in the message are other entities in the message,
`self describing' messages.
We are interested in the size of such messages.

Let the message include the
relationship of each node to all the other nodes. 
Setting $A_x = A_d$ in equation \ref{landmark-deep}
 and assuming $A_d/F < D/2$ and $A_d \ll \log(N)$,
we get:

\begin{equation}
\label{self-describing-big}
D = \frac{\log(N)}{F}
\end{equation}

All messages with at least as many nodes as given by equation 
\ref{self-describing-big}, that  include all the relations between the
nodes, are self describing.

Comparing eq. \ref{landmark-basic} 
to eq. \ref{short-deep} we see that while most sets of $A_x/F$ 
nodes do not have a unique set of relations, about $O(1/\log(N))$ 
of them do. Setting $b = D/2$ and $A_x = A_d = A$ in equation \ref{da},
and approximating, we get the minimum size of a self describing message:

\begin{equation}
\label{self-describing-small}
D = \frac{2A}{F}
\end{equation}

\subsection{Communication Overhead}

 Descriptions can be used to overcome linguistic ambiguity, but at the cost
of added computational complexity in encoding and decoding descriptions.
We now look at the impact of descriptions on the channel capacity required
to send the message. We only consider the case where the sender and
receiver share the same view of the graph.

Consider a message that includes some number of arcs connecting $W$ nodes. If we
assume that the number of distinct arc labels is
much less than $N$, most of the communication cost is in referring to
the $W$ nodes in the message. 

Now, consider the case where each node in the graph is assigned a
unique name. Each name will require $\log(N)$ bits to encode. 
If the message is a random selection of
arcs from graph, we need $W \log(N)$ bits to encode references to the
nodes.

Next consider encoding references to these $W$ nodes using flat
descriptions with unambiguous landmarks where the descriptions are
constructed by randomly picking statements about each node, i.e., $F =
H_g$. This is the case covered by equation \ref{landmark-random-simple}.
We will need descriptions of
length $2 \log(N)/H_g$. These descriptions are strings from the adjacency
matrix and have an entropy rate of $H_g$. So, the string of length 
$2\log(N)/H_g$ can be communicated using $2\log(N)$ bits and references
to $W$ nodes will require $2M\log(N)$ bits. Comparing this to using
unique names for each node, we see that there is a overhead factor of 2.

Now consider encoding references to these $W$ nodes using purely structural
descriptions. These descriptions require $2\log(N)/H_g$ nodes and are of
length $2(\log(N)/H_g)^2$ and we will require $2\log(N)^2/H_g$ bits to
represent each description, giving us an overhead of $2\log(N)/H_g$. We 
see that purely structural descriptions are not only very hard to
decode, but they also incur a significant channel capacity overhead.
This is assuming that the nodes in the message are randomly chosen.
However, if the message is self describing, each of the nodes in the message serves as
descriptors for the other nodes in the message, amortizing the
description cost. Since there are $2\log(N)/H_g$ nodes in the message, 
there is no overhead.




\subsection{Non-identifying descriptions}

 We are  interested in the question of how much can be revealed about
an entity without uniquely identifying it. This is of use in applications involving
sharing data for research purposes, for delivering personalized
content, personalized ads, etc.

We are interested in comparing the number of statements
that can be made about an entity, while still keeping it
indistinguishable from $K$ other entities \cite{kanonymity},
with the number of statements required to uniquely identify it.
For this comparison to be meaningful, the descriptions need to be drawn
from the same ensemble of descriptions.
For a given salience rate $F$, we are interested in how big $D$ can be,
such that at least $K$ other nodes satisfy this description.
Since the purpose is to hide the identity of the entity,
the name of the entity is not included in the
description, i.e., $A_x = \log(N)$.


















Given $N$ nodes and $R$ descriptions, assume that each node is
randomly assigned a description.  The probability
that a given description corresponds to $r$ nodes is approximately:

\begin{equation}
p_r \approx \frac{e^{-\lambda} \lambda^r}{r!}
\end{equation}

\noindent where $\lambda = \frac{N}{R}$, since this is a Poisson process with
parameter $\lambda$.
However, since there are $2^{DA_d}$ interpretations for $D$, let us find the
number of other nodes that also map to these $2^{DA_d}$ descriptions.
Since the sum of Poisson processes is a Poisson process, we find that the probability
of $r$ additional nodes mapping to any of these $2^{DA_d}$ descriptions is

\begin{equation}
p_r \approx \frac{e^{-\lambda 2^{DA_d}} (\lambda 2^{DA_d})^r}{r!}
\end{equation}

\noindent Thus the approximate number of nodes with fewer that $K$ such conflicts
is 
\begin{equation}
\sum_{r=0}^{K-1} Np_r = N e^{-\lambda 2^{DA_d}} \sum_{r=0}^{K-1}\frac{(\lambda 2^{DA_d})^r}{r!}
\end{equation}

\noindent   We want this number to be a small constant $U$, thus
\begin{equation}
\sum_{r=0}^{K-1} \frac{(\lambda 2^{DA_d})^r}{r!} = \frac{U e^{\lambda
    2^{DA_d}}}{N}
\end{equation}

\noindent The left hand side consists of the first $K$ terms of the Taylor
series of $e^x$.  Using Taylor's theorem we have
\begin{equation}
e^x - \sum_{r = 0}^{K - 1} \frac{x^r}{r!}  \leq \frac{e^x x^K}{K!}
\end{equation}
\noindent Substituting the summation, we get:
\begin{equation}
e^{\lambda 2^{DA_d}} -  \frac{Ue^{\lambda 2^{DA_d}}}{N} \leq
\frac{e^{\lambda 2^{DA_d}} {(\lambda 2^{DA_d}})^K}{K!}
\end{equation}
\noindent Thus
\begin{equation}
K!\biggl(1 -  \frac{U}{N}\biggr) \leq (\lambda 2^{DA_d})^K
\end{equation}
\noindent Using Stirling's approximation $K! \approx (K/e)^K$ and taking $K$-th
roots of both sides,
\begin{equation}
\frac{K}{e}\biggl(1 -  \frac{U}{N}\biggr)^{\frac{1}{K}} \leq \lambda 2^{DA_d}
\end{equation}
\noindent Since $U/N$ is small, we can use the binomial approximation to get

\begin{equation}
\frac{K}{e}\biggl(1 -  \frac{U}{KN}\biggr) = \frac{KN - U}{eN} \leq \frac{N 2^{DA_d}}{F}
\end{equation}
\noindent Thus $R \leq \frac{eN^22^{DA_d}}{KN - U}$.  Since $U \ll
KN$, we can ignore $U$. For sufficiently large descriptions, on
average, the probability of a description size $D$ holding for an
object is $2^{-DF}$ where $F$ is the salience rate of the description.
So, we have:

\begin{equation}
 R = 2^{DF} \leq \frac{eN 2^{DA_d}}{K}
\end{equation}

\noindent   Taking logs we get:
\begin{equation}
\label{c-anon}
D \leq \frac{\log N - \log \frac{K}{e}}{F - A_d} \approx \frac{\log N - \log K}{F - A_d}
\end{equation}

\subsubsection{Discussion on non-identifying descriptions}
Comparing equation \ref{c-anon} to equation \ref{adaptive-flat-anon}, we see that $D$ for `$K$-anonymity' 
is very close to the $D$ for the shortest description of that node.
For any node, there are a lot -- ${N \choose D}$ -- of descriptions of size 
$\frac{\log N - \log K}{F - A_d}$. All of
these are satisfied by at least $K$ other nodes. In fact, we could use a slightly 
larger value of $D$, as given in equation \ref{c-anon} and most descriptions
of this size would be satisfied by at least one other node.
Most descriptions of size $\log N/(F - A_d)$ do not
reveal identity. But for every node, there is at least one description
of this size that uniquely identifies it. 
Once the description size exceeds $2 \log N/(F - A_d)$, then, with 
high probability, every description of that size uniquely identifies the node.

This behavior of descriptions --- until a certain size $D < \log N/(F - A_d)$
they are, with high probability ambiguous, but once they cross that threshold,
there is a `phase transition' and 
their ambiguity cannot be guaranteed --- follows from from the
analysis in the earlier section on the phase transition in the likelihood
of finding unique description of size $D$.

We also note that the limits we have derived are for the average
length of descriptions for the entities in a sufficiently long message.
Therefore, limits derived above are a neccessary, but not sufficient
condition for anonymity. If the average size of the descriptions
in a sufficiently large set of descriptions is higher than  that given
by equation \ref{c-anon}, then, with high probability, anonymity
has not been preserved.

\section{Empirical Validation}

 We validate our results for description length for  on a set of 
random graphs with different structural, connection and naming
properties. We look at three kinds of random graphs:

\begin{enumerate}
\item Erdos-Renyi random graph with $N$ nodes, where the
  probability of two nodes being connected is $p$.  Nodes are divided
  into two categories, the first are assigned names with ambiguity
  $A_x$ and the second are assigned names with ambiguity $A_d$.

\item A random bipartite graph where the probability of a node
from one size (same number of nodes on both sides) being connected to a node from the other side is $p$.
Nodes on one side are assigned names with ambiguity $A_x$ and nodes
on the other side are assigned names with ambiguity $A_d$.

\item A random graph with local clustering, constructed as follows: We
  start with a number of Erdos-Renyi graphs that are not connected to
each other. With then pick a number of pairs of nodes from different
clusters and add a link between them, with probability $p$. Within
each node, we divide nodes into two categories and assign them names
with ambiguity $A_x$ and $A_d$.

\end{enumerate}

We look at three categories of descriptions:

\begin{enumerate}
\item Flat descriptions with  unambiguous descriptors (equation \ref{simple}).
\item Flat descriptions  with ambiguous descriptors (equation \ref{adaptive-flat}).
\item Deep descriptions (equation \ref{da}).
\end{enumerate}

We vary the following parameters:

\begin{enumerate}
\item The salience rate for the graph. Note that the salience.
rate for each of these graphs is $-log(p)$.
\item $A_x$, the ambiguity in the nodes being described.
\item $A_x$, the ambiguity in the descriptor nodes
\end{enumerate}

For each value of $p$, $A_x$ and $A_d$, we generated 10 random instances
of the graphs and (with $N =
1000$). For each instance, computed  the shortest description
for 100 of the nodes. We compared the lengths of these descriptions
with those predicted by the corresponding equations. 
Figures 2-9 compare the actual description lengths with 
those predicted by our theory. 

Overall the predicted lengths correspond
fairly closely to the observed lengths. The differences between
observed and predicted lengths are because of the approximations
made in deriving equations \ref{simple}, \ref{adaptive-flat} and
\ref{da}.

\begin{figure}
\centering
\includegraphics[]{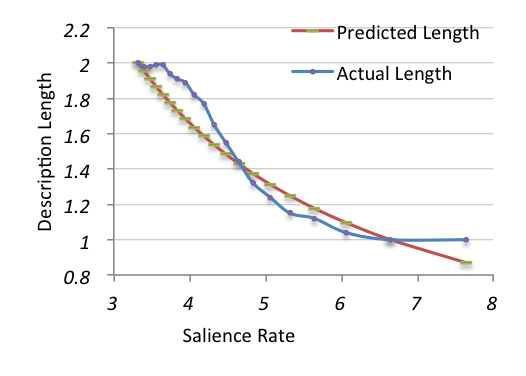}
\caption{Description length as a function of salience, for flat 
descriptions with no ambiguity in descriptor nodes. Ambiguity in
nodes being described is held constant at $\log_2 100$ (i.e., 100
nodes corresponding to each name).}
\end{figure}

\begin{figure}
\centering
\includegraphics[]{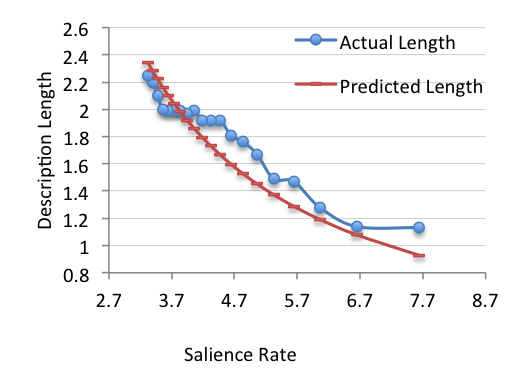}
\caption{Description length as a function of salience, for flat 
descriptions. Ambiguity in
nodes being described is held constant at $\log_2 100$ (i.e., 100
nodes corresponding to each name). Ambiguity in descriptor nodes
is $\log_2 1.4$ (i.e., 1.4 nodes corresponding to each name, on average).}

\end{figure}

\begin{figure}
\centering
\includegraphics[]{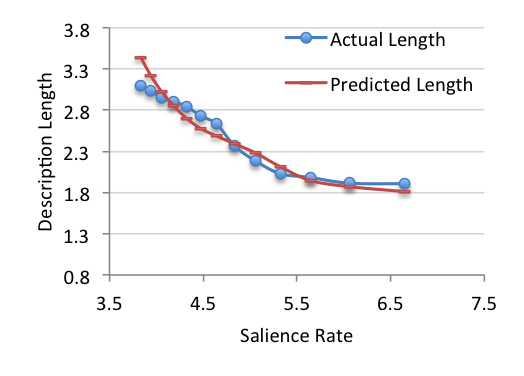}
\caption{Description length as a function of salience, for deep
descriptions. Ambiguity in
nodes being described is held constant at $\log_2 100$ (i.e., 100
nodes corresponding to each name). Ambiguity in descriptor nodes
is $\log_2 8$ (i.e., 8 nodes corresponding to each name, on average).
Number of arcs between descriptor nodes ($bD$) comes from the actual
graph.}

\end{figure}

\begin{figure}
\centering
\includegraphics[]{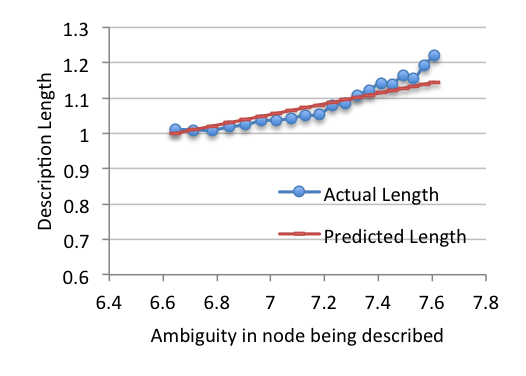}
\caption{Description length as a function of ambiguity in nodes being
described, for flat 
descriptions. Salience rate is held constant at $-\log_2 0.01$ and
there is no ambiguity in the descriptor nodes.}

\end{figure}

\begin{figure}
\centering
\includegraphics[]{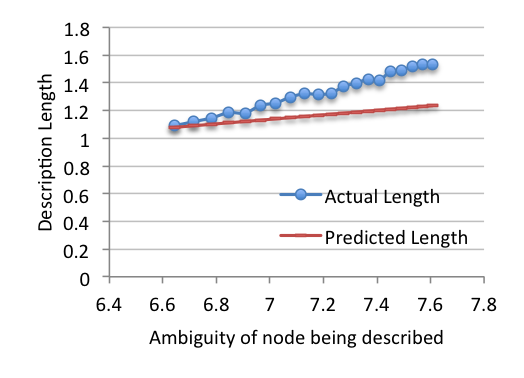}
\caption{Description length as a function of ambiguity in nodes being
described, for flat 
descriptions. Salience rate is held constant at $-\log_2 0.01$.
Ambiguity in descriptor nodes
is $\log_2 1.4$ (i.e., 1.4 nodes corresponding to each name, on average).}

\end{figure}

\begin{figure}
\centering
\includegraphics[]{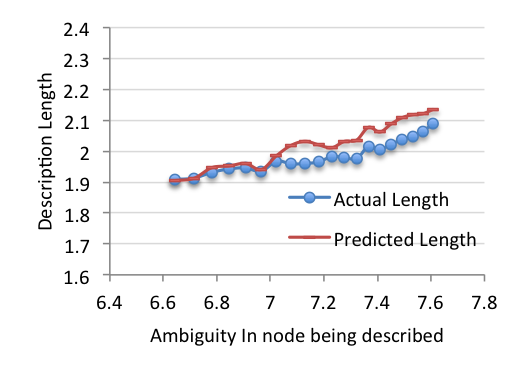}
\caption{Description length as a function of ambiguity in the
nodes being described, for deep
descriptions. Salience rate is held constant at $-\log_2 0.01$.
Ambiguity in descriptor nodes
is $\log_2 10$ (i.e., 10 nodes corresponding to each name, on average).
Number of arcs between descriptor nodes ($bD$) comes from the actual
graph, which accounts for the jagged nature of the predicted length.}
\end{figure}

\begin{figure}
\centering
\includegraphics[]{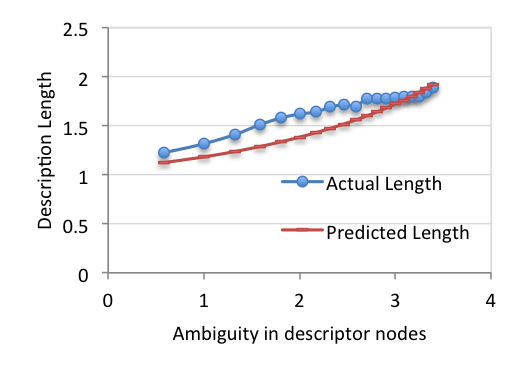}
\caption{Description length as a function of ambiguity in the descriptor
nodes, for flat 
descriptions. Salience rate is held constant at $-\log_2 0.01$.
Ambiguity in the nodes being described is held constant at
$\log_2 100$ (i.e., 100 nodes corresponding to each name, on average).}

\end{figure}

\begin{figure}
\centering
\includegraphics[]{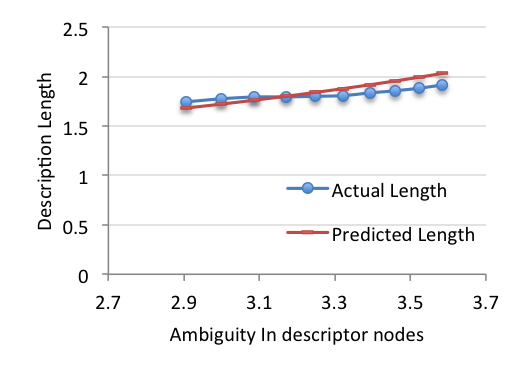}
\caption{Description length as a function of ambiguity in the descriptor
nodes, for deep 
descriptions. Salience rate is held constant at $-\log_2 0.01$.
Ambiguity in the nodes being described is held constant at
$\log_2 100$ (i.e., 100 nodes corresponding to each name, on average).
Number of arcs between descriptor nodes ($bD$) comes from the actual
graph.}

\end{figure}

\section{Ubiquity of Reference by Description}
Reference by description is ubiquitous in everyday human 
communication.
Below  are the results of an analysis of 50 articles from 7 different news
sources, covering 3 different
kinds of articles --- analysis/opinion pieces, breaking news and
wedding announcements/obituaries. We extracted the references to 
people, places and organizations from these articles. In each article entity pair, we
examined the first reference in the article to that entity and
analyzed it. The number of entities referenced and the size
of descriptions, as measured by the number of descriptor entities
in the description, are given in tables 1 and 2.
We found the following:

\begin{enumerate}

\item Almost all references to people have descriptions. The only exceptions
are very well known figures (e.g., Obama, Ronald Reagan). 

\item References to many places, especially countries and large cities, do
  not have associated descriptions. References to smaller places, such 
as smaller cities and neighbourhoods follow a stylized convention,
which gives the city, state and if neccessary, the country.

\item News articles tend to contain more references to
public figures, who have shorter descriptions, reflecting the
assumption of their being known to readers. 

\item Wedding/obituary announcements, 
in contrast, tend to feature descriptions of greater detail. 

\item The
names of many organizations are, in effect, descriptions (e.g., Palo
Alto Unified School District).  
\end{enumerate}

\begin{table}
\label{table2}
\centering
\begin{tabu}{| l |[2pt] c | c | c |[2pt] c |}
\hline
Article Type $\rightarrow$ & Analysis/Opinion & Breaking & Obituaries & Total \\
Source $\downarrow$        & Piece & News & Weddings& \\
\tabucline[2pt]{-}
NY Times &  23  &  6  &  24  &  53 \\
\hline
BBC  & 14  &  24  &  4  &  42 \\
\hline
Atlantic & 121  & 0& 0& 121 \\
\hline 
CNN &  10  &  11  &  6  &  27  \\
\hline 
Telegraph &  12  &  20  &  10 &  42 \\
\hline 
LA Times &  16  &  21  &  9  &  46 \\
\hline
Washing. Post &  28 &  23  &  8  &  59 \\
\tabucline[2pt]{-}
Total &  224  &  104  &  61  &  389 \\
\hline
\end{tabu}
\caption{Number of entity reference in each source, broken down by article type}
\end{table}

\begin{table}
\label{table3}
\centering
\begin{tabu}{| l |[2pt] c | c | c |[2pt] c |}
\hline
Article Type $\rightarrow$ & Analysis/Opinion & Breaking & Obituaries & Average \\
Source $\downarrow$        & Piece & News & Weddings& \\
\tabucline[2pt]{-}
NY Times & 2.9 & 3 & 3.5 & 3.18\\
\hline
BBC  & 3 & 2.54 & 3.5 & 2.78\\
\hline
Atlantic & 3.7 & 0& 0& 3.7\\
\hline 
CNN & 2.5 & 2.63 & 4 & 2.88 \\
\hline 
Telegraph & 2.6 & 2.95 & 3.2 & 2.9\\
\hline 
LA Times & 2.43 & 2.52 & 3.55 & 2.69\\
\hline
Washing. Post & 3.14 & 3.65 & 4 & 3.45\\
\tabucline[2pt]{-}
Average & 3.29 & 2.92 & 3.57 & 3.23\\
\hline
\end{tabu}
\caption{Average description size in each source, broken down by article type}
\end{table}

Given that these articles are stories, the entities that are mentioned
in them are closely related.  Consequently, there exists no  
clear distinction between the parts of
the description that serve to identify an entity from the message itself.

\section{Alternate Problem Formulations }

Descriptions based on random graph models are only one way of
looking at the problem of Reference by Description.
In this section, we briefly look at three alternate
formulations. In all of these formulations, we continue to use
the model described in section 3. We modify our analysis
of descriptions and/or the richness of descriptions.

\subsection{Combinatorial analysis}

Here we look at descriptions from a combinatorial perspective.
Variations exist for the following combinatorial
decision problem: Given a graph with $N$ nodes, $B$ names and description size $D$, 
where each node is assigned one or zero of these names, does each node have
a  unique description, with fewer than $D$ nodes?
In the worst case, $B = 0$, in which case verification of a possible solution
includes solving a subgraph isomorphism problem. Since subgraph
isomorphism is known to be $NP$ complete, this problem is $NP$ complete.
Since there are  ${N \choose D}$ sets of $D$ nodes, we might need to solve 
as many subgraph isomorphism problems.

\subsection{Descriptions and Logical Formulae}

Here, we continue to model the domain of discourse as a directed labelled
graph, but instead of restricting descriptions to subgraphs, we allow for
a richer description language. More specifically, we use formulae in first
order logic as descriptions.

Consider the class of descriptions where the node being described
has no name ($A_x = \log(N)$), where some of the nodes in the description
similarly have no name and others have no ambiguity. This class
of descriptions can be represented as a first 
order formula with a single
free variable  (corresponding to the node being described) and some
constant symbols (nodes
with shared names). The formula is a unique descriptor for a node when the node is
the  only binding for the free variable that satisfies the formula.
The simplest class of descriptions, `flat descriptions', 
corresponds to the logical formula:

\begin{equation}
L_{x_1}(X, S_1) \wedge L_{x_2}(X, S_2) \wedge ... \wedge L_{x_S}(X,S_S)
\end{equation}

\noindent where $L_{x_i}$ is the label of the arc between $X$ and the
$i^{th}$ descriptor node.

Deep descriptions can be seen as introducing existentially
quantified variables (corresponding to the descriptor nodes with no names)
into the description. For the sake of simplicity,
we consider graphs with a single label, $L$ (and $L_{null}$ indicating
no arc). Consider a 
description fragment such as $L_i(X, S_j)$, where $L_i$ is either $L$
or $L_{null}$. Let us allow a single nameless descriptor 
node. This corresponds to 

\begin{equation*}
(\exists y\ L_i(X, y) \wedge L_i(y, S_j)) 
\end{equation*}

Introducing two nameless descriptor nodes corresponds to
\begin{eqnarray*}
(\exists & (y_1, y_2) &\ L_i(X, y_1) \wedge L_i(X, y_2) \\
&\wedge & L_i(y_1, y_2) \wedge L_i(y_2, y_1) \\
& \wedge & L_i(y_1, S_j) \wedge L_i(y_2, S_j))
\end{eqnarray*}

We can define different categories of descriptions by introducing
constraints on the scope of the existential quantifiers. For example
the nameless descriptor nodes can be segregated so that there are separate
subgraphs relating the node being
described to each of the shared nodes. The logical
form of these descriptions (for a single nameless descriptor node) look like: 
\begin{eqnarray*}
& &(\exists y L_i(X, y) \wedge L_i(y, S_1))  \\
&\wedge & (\exists y L_i(X, y) \wedge L_i(y, S_2)) \\
&\wedge &(\exists y L_i(X, y) \wedge L_i(y, S_3)) \wedge \ldots
\end{eqnarray*}

More complex descriptions can be created by allowing universally
quantified variables, disjunctions, negations, etc. The axiomatic
formulation also allows some of the shared domain knowledge to be
expressed as axioms. The down side of such a flexible framework is
that very few guarantees can be made about the computational
complexity of decoding descriptions.

\subsection{Algorithmic descriptions}

 We can allow descriptions to be arbitrary programs that take an entity 
(using some appropriate identifier that cannot be used in the communication)
from the sender and output an entity (using a different identifier, understood
by the receiver) to the receiver.

 This approach allows us to handle graphs with structures/regularities that 
cannot be modeled using stochastic methods. 
An interesting question is the size of the smallest program
required for a given domain, sender and receiver.
If the shared domain knowledge is very low, the program will simply have
to store a mapping from sender identifier to receiver identifier. As
the shared domain knowledge increases, the program can construct 
and interpret descriptions.

\end{document}

%% file: description_figs.tex
\begin{figure}
\ifx\JPicScale\undefined\def\JPicScale{0.9 }\fi
\unitlength \JPicScale mm
\begin{picture} (130,40)(-10,-10)
\thicklines

\put(10,0){\circle{2}}
\put(10,0){\color{green}\circle*{2}}

\put(9,1){\vector(-1,2){11}}
\put(15.5,12){\makebox(0,0)[bl]{\scriptsize $L_3$}}

\put(-2.5,24){\circle{2}}
\put(10,1){\vector(0,1){22}}
\put(5.3,12){\makebox(0,0)[bl]{\scriptsize $L_2$}}
\put(10,24){\circle{2}}
\put(10.5,1){\vector(1, 2.6){8.5}}
\put(-4,12){\makebox(0,0)[bl]{\scriptsize $L_1$}}
\put(19.5,24){\circle{2}}
\put(6, -7) {\makebox(0,0)[bl]{\scriptsize Flat}}

\put(50,0){\circle{2}}
\put(50,0){\color{green}\circle*{2}}
\put(50,1){\line(-1,1){10}}
\put(40,12){\circle{2}}
\put(50,1){\line(0,1){10}}
\put(50,12){\circle{2}}
\put(50,1){\line(1,1){10}}
\put(60,12){\circle{2}}

\put(40,13){\line(1,2){5}}
\put(50,13){\line(-1,2){5}}
\put(45,24){\circle{2}}
\put(40,13){\line(1.5,1){15}}
\put(50,13){\line(1,2){5}}
\put(60,13){\line(-1,2){5}}
\put(55,24){\circle{2}}


\put(45, -7) {\makebox(0,0)[bl]{\scriptsize Deep}}

\put(95,0){\color{green}\circle*{2}}
\put(95,0){\circle{2}}
\put(94.5,1){\line(-1,0.5){10}}
\put(95,1){\line(0,1){7}}
\put(95.5,1){\line(0.5,1){5}}
\put(94,1){\line(-1.5,0.4){18}}

\put(76, 7){\circle{2}}
\put(75.5,8){\line(-1,1.6){5}}
\put(76.5,8){\line(1,1.6){5}}
\put(70.5, 17){\circle{2}}
\put(70.5,18){\line(1,1){5}}
\put(71.5, 17) {\line(1,0){9}}
\put(81.5, 17){\circle{2}}
\put(81.5,18){\line(-1,1){5}}
\put(76,24){\circle{2}}

\put(84, 7){\circle{2}}
\put(84, 8){\line(1,7){1.15}}
\put(95,9){\circle{2}}
\put(94, 9){\line(-1,-0.2){9}}
\put(95,10){\line(-1,0.6){10}}
\put(85,17){\circle{2}}
\put(85,18){\line(0.6,1){3.2}}
\put(89,24){\circle{2}}

\put(100.5,12){\circle{2}}
\put(100.5,13){\line(0,1){5}}
\put(101.5,12){\line(1,0){5}}
\put(100.5,19){\circle{2}}
\put(101.5,20){\line(1,0.5){6}}
\put(107.5,12){\circle{2}}
\put(107.5,13){\line(0,1){10}}
\put(107.5,24){\circle{2}}

\put(85, -7) {\makebox(0,0)[bl]{\scriptsize Intermediate}}
\end{picture}
\caption{Description Shapes}
\end{figure}

%% file: ReferenceByDescription.bbl
\begin{thebibliography}{10}

\bibitem{dimensionality}
C.~C. Aggarwal.
\newblock On k-anonymity and the curse of dimensionality.
\newblock In {\em Proceedings of the 31st international conference on Very
  Large Data Bases}, pages 901--909, 2005.

\bibitem{dbpedia}
C.~Bizer, J.~Lehmann, G.~Kobilarov, S.~Auer, C.~Becker, R.~Cyganiak, and
  S.~Hellmann.
\newblock Dbpedia-a crystallization point for the web of data.
\newblock {\em Web Semantics: science, services and agents on the world wide
  web}, 7(3):154--165, 2009.

\bibitem{freebase}
K.~Bollacker, C.~Evans, P.~Paritosh, T.~Sturge, and J.~Taylor.
\newblock Freebase: a collaboratively created graph database for structuring
  human knowledge.
\newblock In {\em 2008 ACM SIGMOD}, pages 1247--1250. ACM, 2008.

\bibitem{Cohen}
W.~W. Cohen, H.~Kautz, and D.~McAllester.
\newblock Hardening soft information sources.
\newblock In {\em Proceedings of the sixth ACM international conference on
  Knowledge Discovery and Data mining}, pages 255--259, 2000.

\bibitem{Cover}
T.~Cover and J.~Thomas.
\newblock {\em Elements of Information Theory}.
\newblock Wiley-Interscience, 1991.

\bibitem{database}
C.~Data.
\newblock {\em An introduction to database systems}.
\newblock Addison-Wesley publ., 1975.

\bibitem{kanonymity}
D.~Dobkin, A.~K. Jones, and R.~J. Lipton.
\newblock Secure databases: Protection against user influence.
\newblock {\em ACM Transactions on Database systems (TODS)}, 4(1):97--106,
  1979.

\bibitem{Elmagarmid}
A.~K. Elmagarmid, P.~G. Ipeirotis, and V.~S. Verykios.
\newblock Duplicate record detection: A survey.
\newblock {\em IEEE Transactions on Knowledge and Data Engineering}, 19:1--16,
  2007.

\bibitem{erdos}
P.~Erd{\H{o}}s and A.~R{\'e}nyi.
\newblock On random graphs.
\newblock {\em Publicationes Mathematicae Debrecen}, 6:290--297, 1959.

\bibitem{Frege}
G.~Frege.
\newblock Sense and reference.
\newblock {\em The philosophical review}, pages 209--230, 1948.

\bibitem{daphne}
L.~Getoor, N.~Friedman, D.~Koller, and A.~Pfeffer.
\newblock Learning probabilistic relational models.
\newblock In {\em Relational data mining}, pages 307--335. Springer, 2001.

\bibitem{getoor2012entity}
L.~Getoor and A.~Machanavajjhala.
\newblock Entity resolution: theory, practice \& open challenges.
\newblock {\em Proceedings of the VLDB Endowment}, 5(12):2018--2019, 2012.

\bibitem{G14}
R.~V. Guha.
\newblock Communicating and resolving entity references.
\newblock {\em CoRR}, abs/1406.6973, 2014.

\bibitem{G15}
R.~V. Guha and V.~Gupta.
\newblock Communicating semantics: Reference by description.
\newblock {\em CoRR}, abs/1511.06341, 2015.

\bibitem{searchlogs}
A.~Korolova, K.~Kenthapadi, N.~Mishra, and A.~Ntoulas.
\newblock Releasing search queries and clicks privately.
\newblock In {\em Proceedings of the 18th international conference on World
  wide web}, pages 171--180. ACM, 2009.

\bibitem{kripke}
S.~A. Kripke.
\newblock {\em Naming and necessity}.
\newblock Springer, 1972.

\bibitem{nytimes_jmc_obit}
J.~Markoff.
\newblock John mccarthy, pioneer in artificial intelligence, dies at 84.
\newblock {\em New York Times}, 2011.

\bibitem{pii}
A.~Narayanan and V.~Shmatikov.
\newblock Myths and fallacies of personally identifiable information.
\newblock {\em Communications of the ACM}, 53(6):24--26, 2010.

\bibitem{newman1999}
M.~E. Newman and D.~J. Watts.
\newblock Scaling and percolation in the small-world network model.
\newblock {\em Physical Review E}, 60(6):7332, 1999.

\bibitem{Pasula}
H.~Pasula, B.~Marthi, B.~Milch, S.~Russell, and I.~Shpitser.
\newblock Identity uncertainty and citation matching.
\newblock In {\em NIPS}. MIT Press, 2003.

\bibitem{quine}
W.~Quine.
\newblock {\em Mathematical logic}.
\newblock Harvard University Press, 1940.

\bibitem{Russell}
B.~Russell.
\newblock On denoting.
\newblock {\em Mind}, pages 479--493, 1905.

\bibitem{Shannon}
C.~Shannon.
\newblock The mathematical theory of communication.
\newblock {\em Bell System Technical Journal}, 27:379--423, 1948.

\bibitem{watts}
D.~J. Watts and S.~H. Strogatz.
\newblock Collective dynamics of ‘small-world’networks.
\newblock {\em nature}, 393(6684):440--442, 1998.

\bibitem{recordlinkage}
W.~E. Winkler.
\newblock The state of record linkage and current research problems.
\newblock In {\em Statistical Research Division, US Census Bureau}. Citeseer,
  1999.

\end{thebibliography}
